\documentclass[10pt,twocolumn,letterpaper]{article}

\usepackage{cvpr}
\usepackage{times}
\usepackage{epsfig}
\usepackage{graphicx}
\usepackage{amsmath}
\usepackage{amssymb}
\usepackage{textcomp}

\usepackage{xspace}
\usepackage{color}
\usepackage{booktabs}
\usepackage{subcaption}
\usepackage{bbm}
\usepackage{microtype}
\usepackage{tikz}
\usetikzlibrary{positioning}
\usetikzlibrary{arrows.meta}
\usetikzlibrary{calc}
\tikzset{}
\usepackage{pgfplots}
\pgfplotsset{width=2.5cm,compat=newest}

\usepackage[T1]{fontenc}
\usepackage[utf8]{inputenc}

\usepackage{hyperref}
\hypersetup{breaklinks=true,bookmarks=false}

\newcommand{\buyu}[1]{{\textcolor{blue}{BUYU: #1}}}

\newlength\tikzfigwidth
\newlength\tikzfigheight

\def\txtBin{\textrm{b}}
\def\txtMc{\textrm{m}}
\def\txtReg{\textrm{c}}

\def\sa{\Theta}
\def\saBin{\sa_{\txtBin}}
\def\saMc{\sa_{\txtMc}}
\def\saReg{\sa_{\txtReg}}
\def\saBinIdx{\sa_{\txtBin{},i}}
\def\saMcIdx{\sa_{\txtMc,i}}
\def\saRegIdx{\sa_{\txtReg,i}}
\def\saBinNum{M^{\txtBin}}
\def\saMcNum{M^{\txtMc}}
\def\saRegNum{M^{\txtReg}}
\def\numSaRegBins{K}

\def\saBinRealIdx{\saBinIdx}
\def\saMcRealIdx{\saMcIdx}
\def\saRegRealIdx{\saRegIdx}

\def\dsReal{D^{\textrm{r}}}

\def\dsRealSize{N^{\textrm{r}}}

\def\bevmap{\mat{x}}
\def\bevmapH{H}
\def\bevmapW{W}
\def\bevmapC{C}

\def\nnFull{f}
\def\nnFeat{g}

\def\nnAttr{h}

\def\lossSym{\mathcal{L}} 

\def\lossSup{\lossSym_{\textrm{sup}}}
\def\lossSupReal{\lossSup^{\textrm{r}}}

\def\lossBCE{\textrm{BCE}}
\def\lossCE{\textrm{CE}}
\def\lossEllOne{\ell_1}

\def\realspace{\mathbb{R}} 

\newcommand{\mat}[1]{\ensuremath{\mathbf{#1}}}
 
\newcommand{\chis}[2][non]{\ensuremath{\chi^2 \ifthenelse{\equal{#1}{non}}{}{ \left(#1,#2\right)}}} 
\newcommand{\Gammaf}[1][non]{\ensuremath{\Gamma\ifthenelse{\equal{#1}{non}}{}{ \left( #1 \right)}}} 

\newcommand{\degree}[1][non]{\ensuremath{\ifthenelse{\equal{#1}{non}}{^\circ}{#1^\circ}}} 
\LetLtxMacro{\oldsqrt}{\sqrt}
\renewcommand{\sqrt}[1][\ ]{%
  \def\DHLindex{#1}\mathpalette\DHLhksqrt}
\def\DHLhksqrt#1#2{%
  \setbox0=\hbox{$#1\oldsqrt[\DHLindex]{#2\,}$}\dimen0=\ht0
  \advance\dimen0-0.2\ht0
  \setbox2=\hbox{\vrule height\ht0 depth -\dimen0}%
  {\box0\lower0.71pt\box2}}

\makeatletter
\renewcommand{\paragraph}{%
  \@startsection{paragraph}{4}%
  {\z@}{0.5ex \@plus 1ex \@minus .2ex}{-1em}%
  {\normalfont\normalsize\bfseries}%
}
\makeatother

\usepackage{enumitem}
\setitemize[0]{leftmargin=10pt,itemsep=0pt,topsep=2pt,parsep=0pt}
\setenumerate[0]{leftmargin=16pt,itemsep=0pt,topsep=2pt,parsep=0pt}

\cvprfinalcopy 

\def\cvprPaperID{5568} 

\ifcvprfinal\fi
\begin{document}

\title{Understanding Road Layout from Videos as a Whole}

\author{Buyu Liu$^{1}$ $\quad$ Bingbing Zhuang$^{1}$$\quad$ Samuel Schulter$^{1}$$\quad$ Pan Ji$^{1}$ $\quad$ Manmohan Chandraker$^{1,2}$ \\
$^1$NEC Laboratories America $\quad$ $^2$UC San Diego
}


\maketitle
\thispagestyle{empty}

\begin{abstract}
In this paper, we address the problem of inferring the layout of complex road scenes from video sequences. To this end, we formulate it as a top-view road attributes prediction problem and our goal is to predict these attributes for each frame both accurately and consistently. In contrast to prior work, we exploit the following three novel aspects: leveraging camera motions in videos, including context cues and incorporating long-term video information. Specifically, we introduce a model that aims to enforce prediction consistency in videos. Our model consists of one LSTM and one Feature Transform Module (FTM). The former implicitly incorporates the consistency constraint with its hidden states, and the latter explicitly takes the camera motion into consideration when aggregating information along videos. Moreover, we propose to incorporate context information by introducing road participants, e.g. objects, into our model. When the entire video sequence is available, our model is also able to encode both local and global cues, e.g. information from both past and future frames. Experiments on two data sets show that: (1) Incorporating either global or contextual cues improves the prediction accuracy and leveraging both gives the best performance. (2) Introducing the LSTM and FTM modules improves the prediction consistency in videos. (3) The proposed method outperforms the SOTA by a large margin.

\end{abstract}


\section{Introduction}
\label{sec:intro}
Understanding 3D properties of road scenes from single or multiple images is an important and challenging task. Semantic segmentation~\cite{sam:RotaBulo18a,sam:Chen18a,sam:Zhao17a}, (monocular) depth estimation~\cite{sam:Godard17a,sam:Laina16a,sam:Xu18a} and road layout estimation~\cite{sam:Guo12a,sam:Schulter18a,sam:Sengupta12a,Wang_2019_CVPR} are some well-explored directions for single image 3D scene understanding. 
Compared to image-based inputs, videos provide the opportunity to exploit more cues such as temporal coherence, dynamics and context \cite{liu2015multiclass}, yet 3D scene understanding in videos is relatively under-explored, especially with long-term inputs. 
This work takes a step towards 3D road scene understanding in videos through a holistic consideration of local, global and consistency cues.

\begin{figure}\centering
  \includegraphics[width=1.09\linewidth, trim = 0mm 30mm 100mm 80mm, clip]{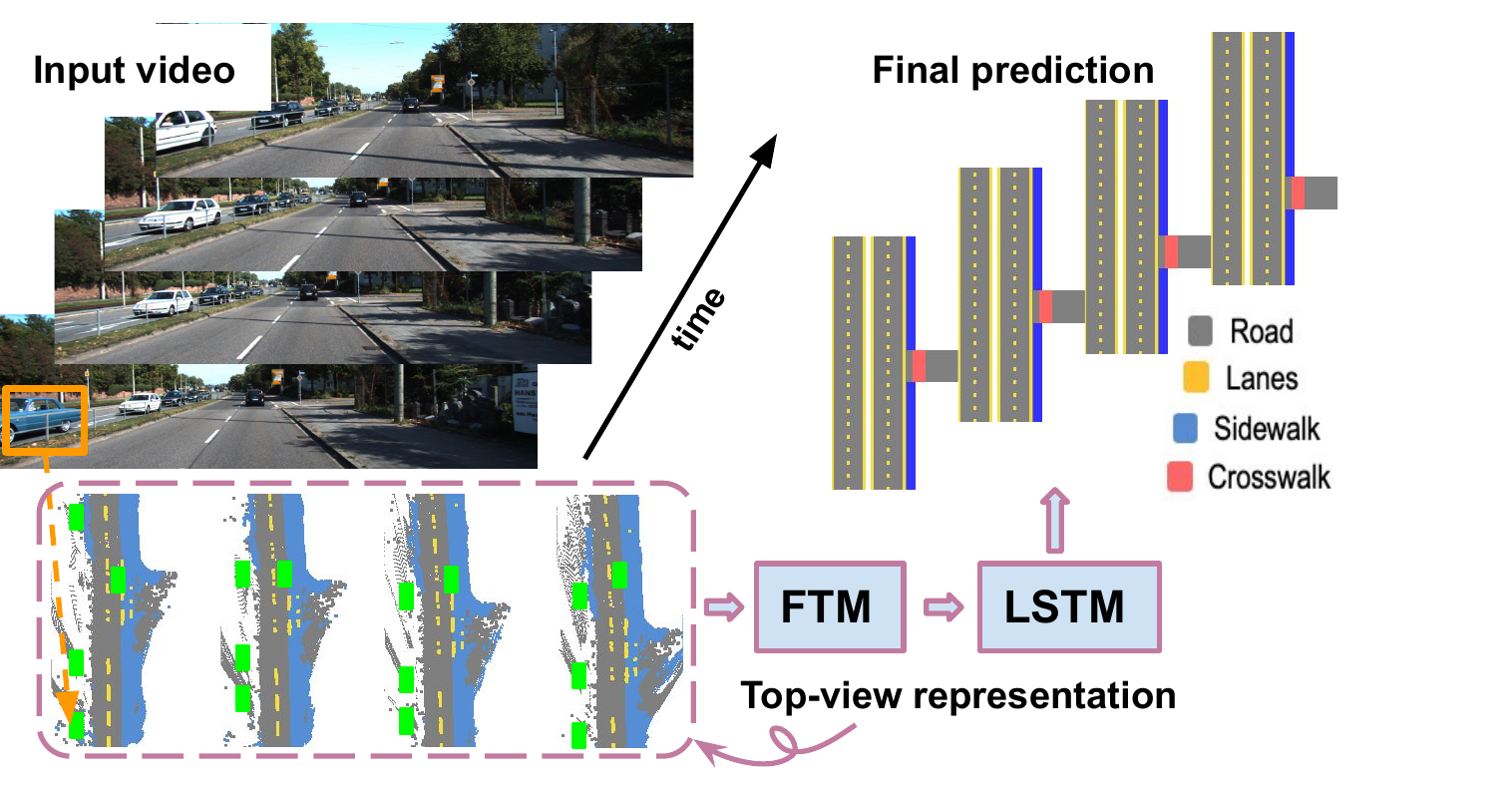}
  \vspace{-0.65cm}
  \caption{
  Given perspective images (top left) that captures a 3D scene, our goal is to predict the layout of complex driving scenes in top-view both  accurately and coherently.
  }
  \label{fig:teaser}
\end{figure}

To this end, we formulate this video scene understanding problem as a road layout estimation problem. We are particularly interested in estimating top-view representations~\cite{sam:Schulter18a,sam:Seff16a} as working in the top-view is beneficial for higher-level reasoning and decision making for downstream applications, e.g. path planning and autonomous driving. 
Given a video sequence, our goal is to predict the road layout of each frame both accurately and coherently. 

To achieve that, we first propose a novel network structure to enforce temporal smoothness in videos. Specifically, our network consists of (i) an LSTM~\cite{hochreiter1997long} that incorporates the long-range prediction coherency and (ii) a Feature Transform Module (FTM) that explicitly aggregates information w.r.t. estimated camera motion and encourages consistency on the feature level. 
By explicitly and implicitly encoding temporal smoothness constraints, our proposed network is able to improve the prediction accuracy and consistency. Apart from incorporating temporal coherency in videos, we further propose to exploit context cues in scene attribute prediction in the top-view. More specifically, such cues are represented and obtained with 3D object detection~\cite{li2019gs3d,li2019stereo} and provide useful priors and constraints for layout estimation in the top-view. For instance, by looking at only the green rectangles in the "top-view representation" in Fig.~\ref{fig:teaser}, one can obtain a rough estimate of the drivable areas.

The above mentioned proposals are applicable for both offline and online scene understanding in videos. When the online scenario is not required, we can further improve our model by leveraging even more powerful temporal cues from future frames. Applications like forensic analysis of traffic scenes and driver behavior studies using commodity cameras are examples where road layout estimation is required while the entire videos can be available. For such cases, we further propose to combine information obtained locally from individual frames with that obtained globally from entire video sequences. Specifically, we utilize structure from motion (SfM) \cite{schonberger2016structure} and multiple view stereo (MVS) \cite{schonberger2016pixelwise} to obtain a dense 3D reconstruction of the road scene from the video sequence. The multiview geometric constraints encoded in such methods naturally aggregate the temporal information from multiple frames, thus permitting a more coherent scene representation as compared to individual views.
This allows us to build the model input that is of better representation power and smoother, which boosts the prediction accuracy as well as coherence (see Fig.~\ref{fig:teaser}).

We conduct experiments to validate our ideas on two public driving data sets, KITTI~\cite{sam:Geiger13a} and NuScenes~\cite{sam:NuScenes18a} (Sec.~\ref{sec:exps}).  Our results demonstrate the effectiveness of the global and context cues in terms of prediction accuracy, and the importance of the LSTM as well as FTM for consistent outputs.
To summarize, our key contributions are:
\begin{itemize}
\item A novel neural network that includes (i) an LSTM that implicitly enforces prediction coherency in videos and (ii) a Feature Transform Module that explicitly encodes the feature level consistency w.r.t. camera motion.
\item An input representation that considers information obtained (i) locally from individual frames, (ii) globally from entire video sequence and (iii) from contextual cues.
\item Experimental evaluation that outperforms state-of-the-art on public datasets by a large margin.
\end{itemize}


\section{Related Work}
\label{sec:related}
Scene understanding is an important yet challenging task in computer vision, which enables us to perform applications such as robot navigation~\cite{sam:Gupta17a}, autonomous driving~\cite{sam:Geiger14a,sam:Kunze18a}, augmented reality~\cite{sam:Armeni16a} or real estate~\cite{sam:Liu15a,sam:Song18a}.

\paragraph{3D scene understanding:} 3D scene understanding 
is most frequently explored for indoor scenes~\cite{sam:Armeni16a,sam:Liu15a,sam:Song17a},  which is typically formulated as a room layout estimation problem. However, unlike the indoor scenes where strong priors, e.g. Manhattan world assumption, are available, scene understanding for outdoor scenarios is less constrained, thus can be more challenging. To this end, many non-parametric approaches have been proposed~\cite{sam:Guo12a,Tighe_2014_CVPR,sam:Tulsiani18a}, where layered representations are utilized to reason about the geometry as well as semantics in occluded areas. 
More recent work~\cite{sam:Schulter18a,sam:Sengupta12a} propose top-view representations to provide a more detailed description for outdoor scenes in 3D. As for parametric models, Seff and Xiao~\cite{sam:Seff16a} propose a model that consists of certain road scene attributes and further utilize a neural network to directly predict them from a single perspective RGB image. However, these pre-defined attributes are not rich enough to capture various types of road layout. A more recent work~\cite{sam:Kunze18a} presents a graph-based road representation that includes lanes and lane markings, from partial segmentation of an image. However, this representation focuses on straight roads only. Similarly, an interesting model is proposed by M\'{a}ttyus~\etal~\cite{sam:Mattyus16a} to augment existing map data with richer semantics. Again, this model only handles straight roads and requires additional input from aerial images. To further handle complex road layouts with traffic participants, Geiger~\etal~\cite{sam:Geiger14a} propose to utilize multiple modalities as input, such as vehicle tracklets, vanishing points and scene flow. Recently, a richer model~\cite{Wang_2019_CVPR} is proposed to handle more complex road scenarios, e.g. multiple lanes and different types of intersections. Our work follows the parametric representation proposed in~\cite{Wang_2019_CVPR}. Unlike~\cite{Wang_2019_CVPR} that only takes local information, e.g. pixel-level depth and semantic segmentation from a single frame, we propose to explore multiple aspects in videos, e.g. global information and context cues, to obtain accurate and coherent predictions.

\vspace{-0.5cm}
\paragraph{Scene understanding in videos:} 
Apart from accuracy, scene understanding in videos further requires consistent predictions between consecutive frames. Spatio-temporal probabilistic graphical models~\cite{liu2015multiclass,Wang_2019_CVPR} are widely explored for outdoor scenes. Nowadays, recurrent neural networks (RNNs) like LSTMs~\cite{donahue2015long,hochreiter1997long,srivastava2015unsupervised} are used to propagate feature representations from still frames over time. However, LSTMs implicitly enforce the prediction consistency without the explicit knowledge of motion. To this end, more recent work~\cite{feichtenhofer2017spatiotemporal,simonyan2014two} propose to combine features extracted from both images and motion, or optical flow, to boost the representational power. Although motion is fed as additional input for networks in these above mentioned networks, it is not utilized to transform features over time to explicitly enforce the feature level consistency. Maybe~\cite{vu2018memory,zhu2017flow} are the most closest work to ours in terms of feature warping and aggregation. Specifically, ~\cite{zhu2017flow,zhu2017deep} explicitly warp feature maps~\cite{fischer2015flownet} between frames and propose to aggregate them in a more well-aligned manner. To address the fixed temporal width problem in~\cite{zhu2017flow}, more recent work~\cite{vu2018memory} introduces a feature memory that is warped from one frame to another, which enables a longer temporal horizon without looking into future frames. In contrast, we introduce a Feature Transform Module (FTM) that warps features w.r.t camera motion and aggregates them between consecutive frames. And we further propose to combine LSTM and FTM to implicitly and explicitly enforce temporal smoothness in predictions. 
More importantly, unlike prior methods that estimate flow in the perspective view of the scene, all our modules work in the top-view space.

\begin{figure*}[t]
 \setlength{\belowcaptionskip}{-0.35cm}
 \centering
 \vspace{-0.3cm}
  \includegraphics[width=1.0\linewidth, trim = 0mm 100mm 0mm 0mm, clip]{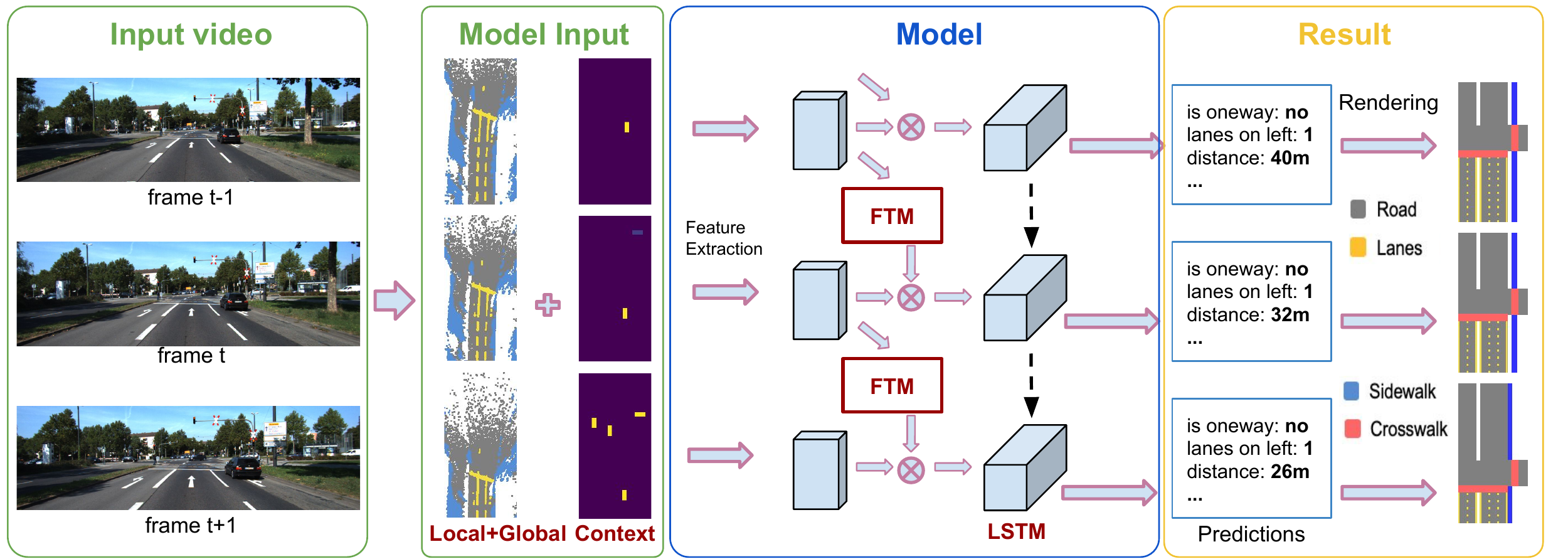}
  \vspace{-0.65cm}
  \caption{\textbf{Overview of our proposed framework:} Given videos as input, top-view maps aggregated from local, global and context information, are fed into our FTM/LSTM network to predict the parametric road scene layout. 
  }
  \label{fig:framework_overview}
\end{figure*}


\section{Our Framework}
\label{sec:method}

We briefly introduce the parameterized scene model in Sec.~\ref{sec:scene_model}. Then we describe our proposed LSTM with FTM in Sec.~\ref{sec:train_infer_scene_model}, followed by the utlization of context information and global information in Sec.~\ref{sec:bev_maps}. An overview of our framework is shown in Fig.~\ref{fig:framework_overview}.


\subsection{Scene Representation}
\label{sec:scene_model}
Given a monocular video sequence capturing a road in the perspective view, our goal is to obtain a coherent yet accurate road layout estimation in the top-view for each frame. To describe the top-view layout, we follow the parametric representation of~\cite{Wang_2019_CVPR}, which consists of three different types of scene attributes/variables, including $\saBinNum = 14$ binary variables $\saBin$, $\saMcNum = 2$ multi-class variables $\saMc$ and $\saRegNum = 22$ continuous variables $\saReg$.
These parameters can represent a large variety of road scene layouts~\footnote{We refer to our supplementary material and~\cite{Wang_2019_CVPR} for more details.}. See Fig.~\ref{fig:teaser} for examples of predicted results.
Denoting the scene model parameters $\sa=\{\saBin,\saMc,\saReg\}$, we aim to predict $\sa^t$, the scene parameters of the $t$-th frame, coherently and accurately for all $t\in\{1,\dots,T\}$, where $T$ is the number of frames in a video. However, instead of making predictions from single image, we propose to augment the underlying feature representation and aggregate temporal information from videos for more temporally consistent and robust predictions.


\subsection{Our LSTM-based Model}
\label{sec:train_infer_scene_model}
In order to exploit long-term temporal information, we propose to utilize LSTM for road layout prediction. In addition, we propose a Feture Transform Module (FTM) to more explicitly aggregate temporal information. We will discuss more details below.

\begin{figure}[t]
	\setlength{\belowcaptionskip}{-0.35cm}
    \begin{center}
        \includegraphics[width=1.8\linewidth, trim = 0mm 40mm 25mm 100mm, clip]{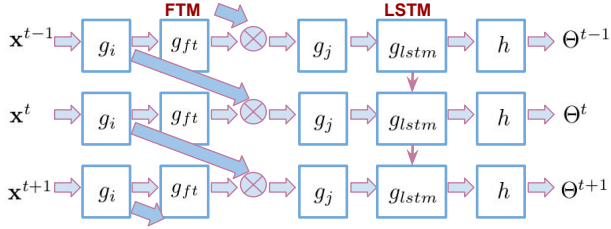}
    \end{center}
    \vspace{-0.75cm}
    \caption{Overview of our network.}
    \label{fig:model_overview}
\end{figure}

We denote the representation of the road scene at a certain view as $\bevmap \in \realspace^{H \times W \times C}$, and hence $\bevmap^t$ at the $t$-th frame in a video. Here, $\bevmap$ can be regarded as any generic representation such as RGB perspective image or top-view image of the road; our proposed novel top-view representation will be discussed shortly in Sec.~\ref{sec:bev_maps}.
Given $\bevmap^t$, our overall model is defined as following:
\vspace{-0.1cm}
\begin{equation}
\begin{split}
    y^t = \nnFull_{com}(\nnFeat_{i}(\bevmap^t),\nnFeat_{ft}(\nnFeat_{i}(\bevmap^{t-1}))), \\
    \sa^t = \nnAttr(\nnFeat_{lstm}( \nnFeat_{j}(y^t))),
\end{split}
\label{eq:total_NN}
\end{equation}
\vspace{-0.1cm}

\noindent where $\nnAttr$, $\nnFeat_{*}$ and $\nnFull_{*}$ are neural networks, with weights 
$\gamma_h$, 
$\gamma_{g_*}$ and $\gamma_{f_*}$
respectively, that we want to train.
$y^t$ is the auxiliary intermediate feature representation.
Our network structure is illustrated in Fig.~\ref{fig:model_overview}.
The architecture of $\nnFeat_i$ is a shallow (one-layer) convolutional neural network (CNN) and $\nnFeat_j$ is a deeper (seven-layer) network. We firstly pass the individual input $\bevmap^t$ to $\nnFeat_i$ and receive feature $\nnFeat_i(\bevmap^t)$. Then this feature is combined with $\nnFeat_{ft}(\nnFeat_i(\bevmap^{t-1}))$, which is obtained by feeding $\nnFeat_i(\bevmap^{t-1})$, the feature obtained from previous frame, to FTM $\nnFeat_{ft}$, and feed to $\nnFeat_j$. The output of $\nnFeat_j$ is a 1-dimensional feature vector (for each frame) that is further fed into our LSTM module $\nnFeat_{lstm}$. $\nnFeat_{lstm}$ then outputs features that implicitly encode information from previous frame by incorporating hidden states from $\bevmap^{t-1}$ and send to $\nnAttr$.
Then, the function $\nnAttr$ is defined as a multi-layer perceptron (MLP) predicting the scene attributes $\sa^t$ with features obtained from LSTM module.
Specifically, $\nnAttr$ is implemented as a multi-task network with three separate predictions for each of the parameter groups of the scene model~\footnote{In experiments, we also tried both bi-directional LSTM~\cite{schuster1997bidirectional} and ConvLSTM~\cite{xingjian2015convolutional}. We did not observe significant improvements over traditional LSTM.}.

\subsubsection{Feature transform module}
\label{sec:model_STM}

Assuming that after $\nnFeat_i$, we are able to extract a feature map $\mathbf{F} = \nnFeat_i(\bevmap) \in \realspace^{h_f \times w_f \times c_f}$, where $h_f$, $w_f$, and $c_f$ denote the height, width and feature dimension. We would like the network to learn more robust feature representations by encouraging temporal consistency. To this end, we propose to aggregate the features that correspond to the same location in the scene from the nearby frames, and feed the composited features into the later prediction layers. Intuitively, we would like the network to make use of the features at the same point but from different views for more robust prediction, as well as encourage more temporally consistent feature learning.


\noindent \textbf{Feature correspondence:}
In order to aggregate the features corresponding to the same spatial location, we need to find a correspondence between two feature maps in neighbouring frames.
One obvious way to achieve this is by computing optical flow~\cite{sun2010secrets} between neighbouring perspective images and transforming the correspondence w.r.t. the flow field on the feature map. We later
show in Sec.~\ref{sec:input_ind_and_colmap} that such computation can be obviated here as we have the camera poses and 3D reconstruction of the road built from the video.

\noindent \textbf{Feature transformation:}
Given the correspondence between feature maps, we can warp~\cite{vu2018memory,zhu2017flow} the feature map at time step $t-1$ to the current time step $t$, denoted as: 
\begin{equation} 
    \nnFeat_{ft}(\mathbf{F}^{t-1}) = \phi(\mathbf{F}^{t-1},\ S^{t-1}(\bevmap^{t},\bevmap^{t-1}))
\label{eq:STM}
\end{equation} 
where $\phi(*)$ is the bilinear sampling function~\cite{sam:Jaderberg15a} and $S^{t-1}(\bevmap^{t},\bevmap^{t-1}) \in \realspace^{h_f \times w_f \times 2}$ is a displacement (or flow) field between frames $t$ and $t$-1.

\noindent \textbf{Feature composition:}
One can choose various ways to aggregate the feature map $\nnFeat_{ft}(\mathbf{F}^{t-1})$ and $\mathbf{F}^{t}$. Here, we adopt the following simple weighted summation of the two, 
\begin{equation}
    \nnFull_{com}(\mathbf{F}^t,\nnFeat_{ft}(\mathbf{F}^{t-1})) = \alpha \cdot \mathbf{F}^t +(1-\alpha)\cdot \nnFeat_{ft}(\mathbf{F}^{t-1}),
\end{equation}
where $\alpha$ can either be a scalar, e.g. $\alpha=\frac{1}{2}$, or a matrix, $\alpha \in \realspace^{h_f \times w_f}$. Note that in both cases, $\alpha$ can be automatically learned together with the rest of the network without any additional supervision. For special cases where $t = 1$, we simply assume $\nnFeat_{st}(\mathbf{F}^{t-1})$ equals to $\mathbf{F}^t$. It's simple but it works well in practice, see supp for more discussion.

\noindent \textbf{Discussion:} Note that we can easily extend our model by aggregating features from frames that are further away or even from future frames in offline settings. Also, FTM can be extended for multiple feature maps at different scales, which can potentially further improve the performance.

\subsubsection{Loss functions on scene attribute annotation:}
Given the annotated data sets $\dsReal$ that consist of $\dsRealSize$ videos, we define our overall loss function $\lossSupReal$ as:
\vspace{-0.2cm}
$$\sum_{i,t}^{\dsRealSize,T} \lossBCE(\saBinRealIdx^{t}, \hat{\Theta}^t_{b,i})  + \lossCE(\saMcRealIdx^t, \hat{\Theta}^t_{m,i})+ \lossEllOne(\saRegRealIdx^t, \hat{\Theta}^t_{c,i}), $$
\vspace{-0.4cm}

\noindent where CE denotes cross-entropy and BCE represents binary cross-entropy. We further denote the ground-truth layout parameters of the $t$-th frame in the $i$-th video sequence in the data set as $\hat{\Theta}^t_{*,i}$. Note that we discretize continuous variables for the regression task -- each variable is discretized into $\numSaRegBins$=100 bins by convolving a Dirac delta function centered at $\saReg$ with a Gaussian of fixed variance~\cite{Wang_2019_CVPR}. 

\subsection{Model Input}
\label{sec:bev_maps}
As suggested by~\cite{Wang_2019_CVPR}, the form of our model input $\bevmap$ has a large impact on our model. Although a perspective RGB image is a natural representation of $\bevmap$, it does not work well in top-view layout estimations. Instead,~\cite{Wang_2019_CVPR} proposes to convert each single perspective image to semantic top-view with pixel-level depth and semantic segmentation. In our case, since we are working on videos, there are more aspects that can be explored, like long-term temporal smoothness. In this section, we propose to exploit both context and global information in videos to improve the representation of $\bevmap$.

\paragraph{Local information: single frame}
Following \cite{Wang_2019_CVPR}, given a single perspective image, together with its semantic segmentation, CNN-predicted dense depth, and camera intrinsics, we could obtain a top-view image of the road by (1) back-projecting all the road pixels into a 3D point cloud and (2) projecting all the 3D points onto the x-y plane.
We use this mapping to transfer the semantic class probability distribution of each pixel from perspective- into top-view. We call it $bev$ and denote it as $\bevmap \in \realspace^{\bevmapH \times \bevmapW \times \bevmapC}$, where $\bevmapC$ is the number of semantic classes. 
$\bevmapH= 128$ and $\bevmapW= 64$ pixels, which relates to 60×30 meters in the point cloud. See Fig.~\ref{fig:bevinit-examples} for some examples of the $bev$ map obtained with single images. 

\subsubsection{Context information}
\label{sec:context}
\begin{figure}
    \setlength{\belowcaptionskip}{-0.35cm}
	\begin{center}
	    \vspace{-0.35cm}
		\includegraphics[width=0.4\textwidth, trim = 8mm 65mm 85mm 0mm, clip]{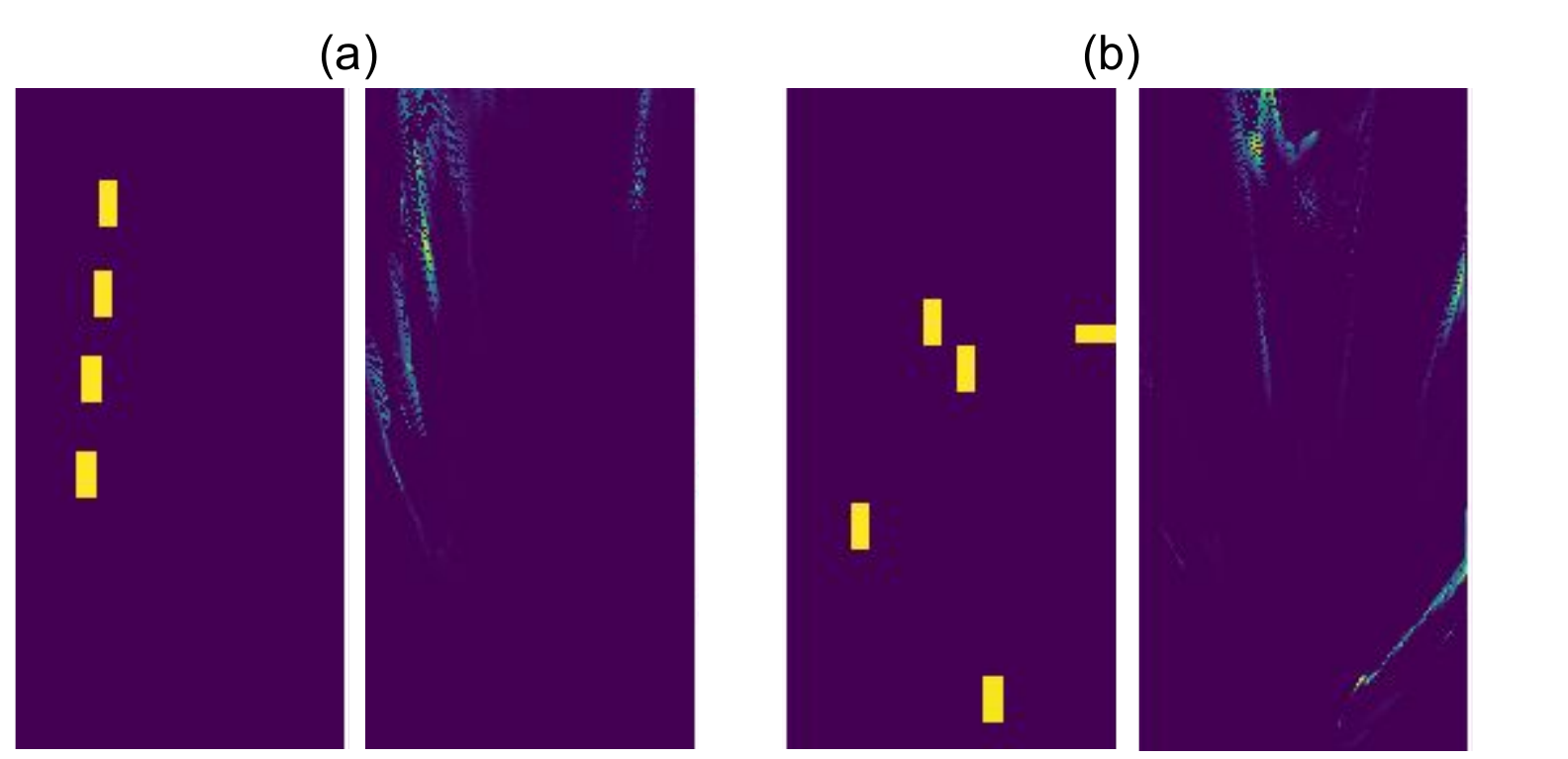} \\
	\end{center}	
	\vspace{-0.55cm}
  \caption{Two examples comparing the box representation and the point cloud representation of the cars. We can observe four frontal-faced cars in both (a) and (b) and one side-faced car in (b). 
  }
  \label{fig:obj_example}
\end{figure}

Considering that incorporating context cues proves to be very beneficial in many scene understanding tasks, unlike \cite{Wang_2019_CVPR} that only includes $C=4$ classes (i.e. road, sidewalk, lane, crosswalk), we propose to further encode an object class in $\bevmap$ (so $C=5$). Specifically, we observe that traffic participants, i.e. vehicles like cars, are commonly present in driving scenarios and can be very useful for layout estimation, e.g. side-faced cars are informative in terms of predicting the existence of or even the distance to side-road. 
One naive way to incorporate context cues
is to follow what is done for other classes to directly map the pixel-level probabilities of objects to top-view. However, we observe that this naive method would lead to diffused representations\footnote{This is mainly due to the small size of the objects, making it less robust to semantic and depth prediction noise when mapping to top-view.} (See Fig.~\ref{fig:obj_example} for examples) and can be very noisy. Instead, we propose to utilize the bounding box representation. Specifically, we apply existing 3D object detectors~\cite{li2019gs3d,li2019stereo} on our videos and map the detection results into top-view. In order to reduce the impact of angle predictions, we further propose to categorize angles into two types, frontal-faced and side-faced. As long as the predicted yaw angle is within $[-\frac{\pi}{4},\frac{\pi}{4}]$ or $[\frac{3\pi}{4},\frac{5\pi}{4}]$ w.r.t. camera $z$-axis (forward direction), we assume this predicted object is frontal-faced. Otherwise, it is side-faced (Fig.~\ref{fig:obj_example}).

\vspace{-0.35cm}

\subsubsection{Global information from video sequence}
\label{sec:input_ind_and_colmap}

\begin{figure}
    \setlength{\belowcaptionskip}{-0.35cm}
	\centering
	   \includegraphics[width=0.4\textwidth, trim = 0mm 73mm 140mm 0mm, clip]{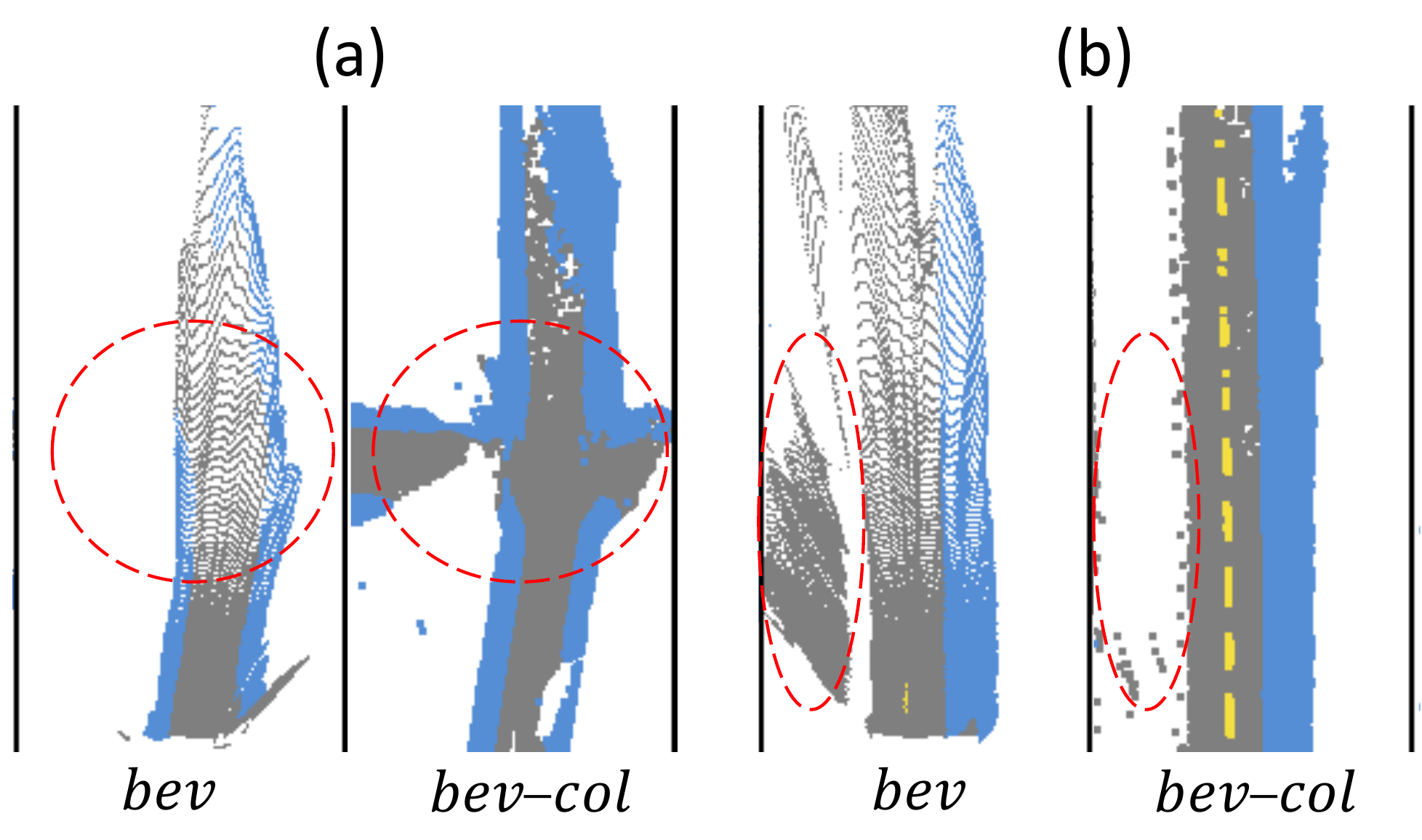}
	\vspace{-0.35cm}
  \caption{Two examples of $bev$ and $bev$-$col$ on the background classes obtained from individual frame and COLMAP. Red circles highlight the differences.}
  \label{fig:bevinit-examples}
\end{figure}

As can be viewed in individual frame examples in Fig.~\ref{fig:bevinit-examples}, there are clearly some artifacts with $bev$, e.g. the V-shape in the bottom and the missing details where pixels are far away. This is in essence due to the inherent limitation of single-view 3D reconstruction in terms of the resolution for far away points in the scene.

\begin{figure}\centering
\setlength{\belowcaptionskip}{-0.35cm}
\vspace{-0.35cm}
  \includegraphics[width=0.4\textwidth, trim = 10mm 0mm 120mm 0mm, clip]{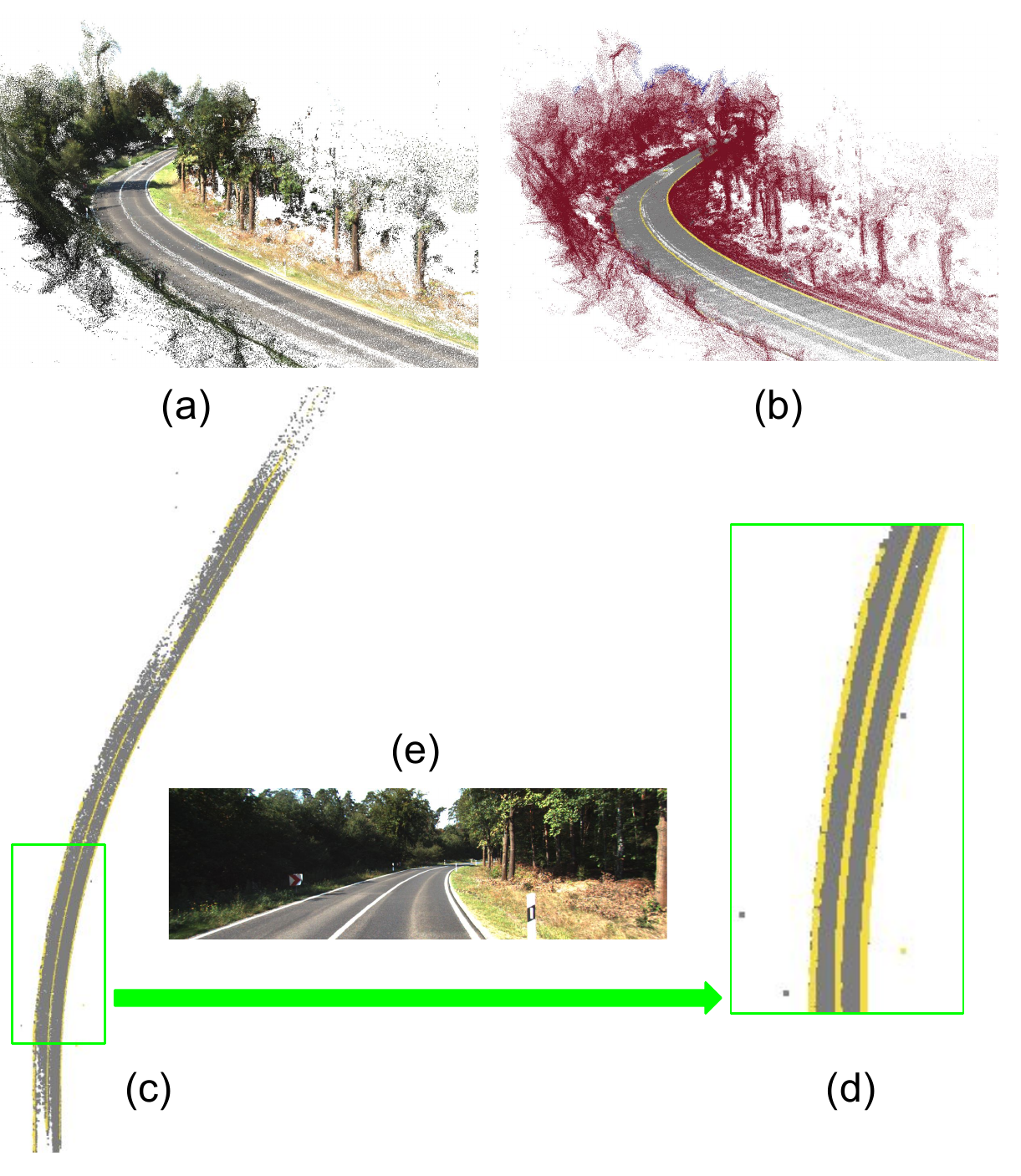}
  \vspace{-0.6cm}
  \caption{An example illustration of generating semantic $bev$-$col$. (a)(b) 3D scene reconstruction and semantic segmentation. (c) road extraction and plane fitting. (d) $bev$-$col$ for the view (e). }
  \label{fig:bevfromcolmap}
\end{figure}

\noindent \textbf{Top-view map from COLMAP:} When the input is an entire video sequence, we are able to address the above mentioned artifacts in an individual frame by generating a more complete 3D reconstruction as well as the subsequent $bev$. Specifically, we make use of state-of-the-art SfM and MVS algorithms from COLMAP~\cite{schonberger2016structure,schonberger2016pixelwise}, as well as the semantic segmentation in 2D image, in order to generate the top-view map (denoted as $bev$-$col$) from a video. The pipeline is illustrated in Fig.~\ref{fig:bevfromcolmap}. We first perform 3D semantic segmentation on the dense point cloud returned from COLAMP. We project each 3D point into the images as per the visibility map to obtain the semantic candidates from the 2D semantic segmentation; a simple winner-take-all strategy is applied to determine the final semantic label for each point (Fig.~\ref{fig:bevfromcolmap}(b)). The 3D semantic segmentation allows us to extract the 3D points for the road part, to which we fit a 2D plane via RANSAC (Fig.~\ref{fig:bevfromcolmap}(c)). 
Finally, we can generate the $bev$-$col$ by cropping a rectangular box on the 2D road plane according to the camera pose and our pre-defined top-view image size and resolution; the 3D points on the plane are converted into the pixels in the image (Fig.~\ref{fig:bevfromcolmap}(d)). After obtaining the global plane parameters, we utilize the camera pose to transform the global plane parameters to the local parameters with respect to the camera. Note that in a well-calibrated case where the global plane can be the same as local plane, the global lane estimation process can be skipped.
See Fig.~\ref{fig:bevinit-examples} for two more examples. Here, one can see that the dense correspondence between two $bev$-$col$ can be trivially obtained given the known position and orientation of each $bev$-$col$ in the 2D road plane. Such mapping is transferred to the correspondence on the feature map and fed into FTM (Sec.~\ref{sec:model_STM}). Here, we do not handle the mapping of dynamic objects, but find FTM to still work well~\footnote{Having access to GT masks for dynamic objects may help to improve performance, but in their absence, even our current handling of dynamic objects yields promising results.}. Finally, we also note that the absolute scale of the 3D reconstruction here can be retrieved from either GPS or camera-ground height in driving scenes.

\noindent \textbf{Discussion:} Note that in our case, rather than presenting what can be actually seen in the current frame, our $bev$-$col$ obtained with COLMAP models what really exists. For instance, a sideroad that is far away or occluded can hardly be noticed in $bev$ generated from an individual frame only due to the lack of observations in perspective view. In contrast, $bev$-$col$ generated with global information is able to recover the sideroad properly as long as we have observations in this video (see comparison in Fig.~\ref{fig:bevinit-examples}(a)).   

\begin{figure}
    \setlength{\belowcaptionskip}{-0.35cm}
	\centering
	\begin{tabular}{c}
		\includegraphics[width=0.45\textwidth, trim = 0mm 65mm 80mm 5mm, clip]{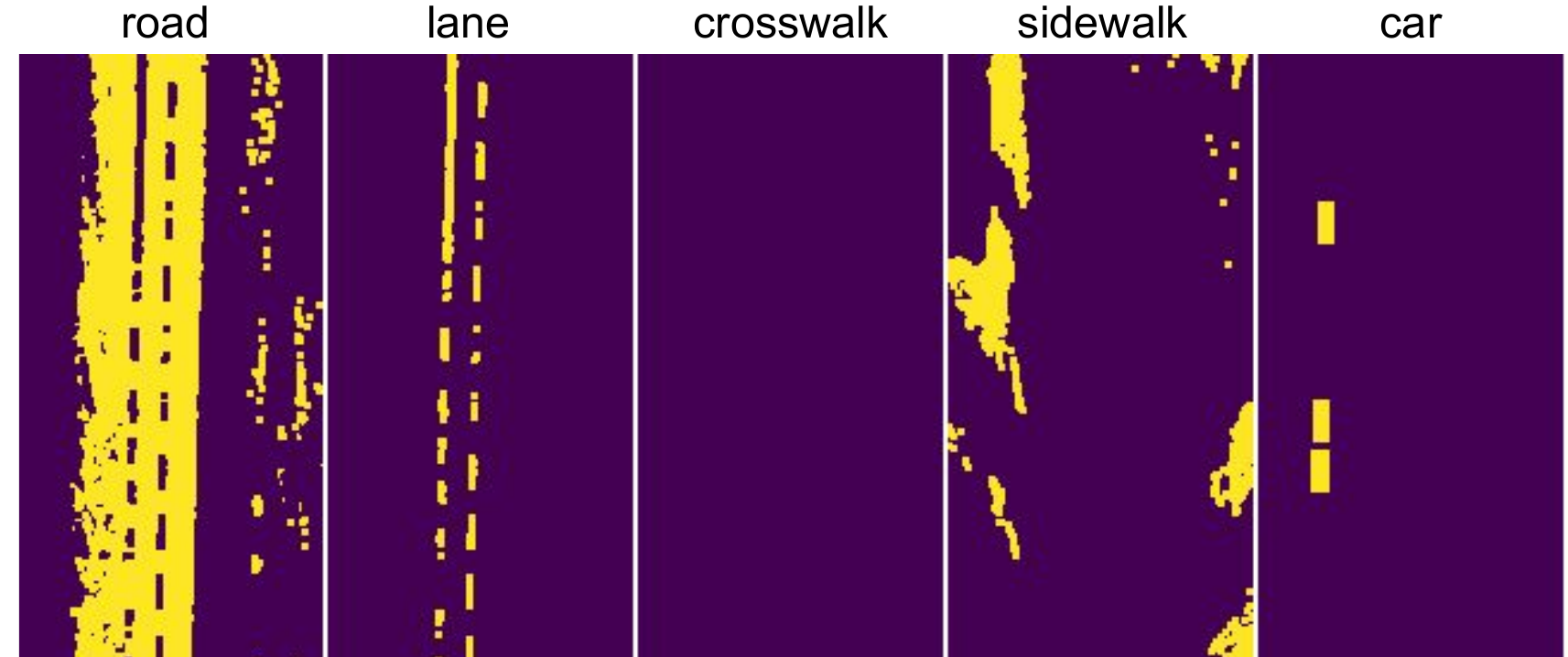} \\
	\end{tabular}
	\vspace{-0.35cm}
  \caption{An example visualizing the layers in our $bev$-$final$. 
  }
  \label{fig:bevinit-context-example}
\end{figure}

\noindent \textbf{Final input:} Our final input incorporates local information from individual frame, global information from videos and object-level context information. Note that local and global cues can be mutually informative, e.g. geometry based 3D reconstruction might be inaccurate or incomplete in some challenging cases like texture-less regions, but in such cases the top-view information from CNN-predicted depth could be more informative towards the layout prediction (See comparison in Fig.~\ref{fig:bevinit-examples}(b)). In summary, our proposed top-view semantic map $bev$-$final$ is in $\realspace^{\bevmapH \times \bevmapW \times \bevmapC}$, where $\bevmapC=5$ and the layers represents four background classes and car; We overlay $bev\text{-}col$ on top of $bev$ to fuse them together. Specifically, for a binary mask $B$ with $B_{i,j}{=}1$ if $bev\text{-}col_{i,j}{\neq}0$, we have $bev \text{-}final=bev\text{-}col{\odot} B+bev{\odot}(1{-}B)$, where $\odot$ indicates element-wise product. See an example visualization of each layer in Fig.~\ref{fig:bevinit-context-example}. 

\section{Experiments}
\label{sec:exps}


In this section, we demonstrate the effectiveness of our proposed augmented input representation and LSTM/FTM by conducting several experiments on two data sets. 

\paragraph{Datasets:}
We validate our ideas on two data sets, KITTI~\cite{sam:Geiger13a} and NuScenes~\cite{sam:NuScenes18a}. Similarly, we utilize the annotated data in~\cite{Wang_2019_CVPR}, which includes around 17000 annotations for KITTI~\cite{sam:Geiger13a} and about 1200 annotations 
for NuScenes~\cite{sam:NuScenes18a} in terms of scene layout annotation. To train our segmentation network, we further have 1243 images for training and 382 for testing on KITTI. We refer the readers to~\cite{Wang_2019_CVPR,Wang_2019_CVPRworkshop} for more details of data sets and implementation details can be found in supplementary. 

\begin{table}\centering\small
\setlength{\belowcaptionskip}{-0.4cm}

\begin{tabular}{l|cccc}
  \hline
              & \multicolumn{4}{c}{NuScenes~\cite{sam:NuScenes18a}} \\
  \hline
  Method      & Accu.-Bi. $\uparrow$ &  Accu.-Mc. $\uparrow$ & MSE $\downarrow$ & IOU $\uparrow$ \\
  \hline\hline
  BEV          & .846    & .485      & .073 & .217\\
  \hline
  BEV-C          & .856    & .471      & .069 & .211\\
  BEV-J        & \textbf{.872}      & .486      & .036 & .230\\
  BEV-J-O      & .858      & \textbf{.543}      & .027 & \textbf{.313}\\  
  \hline
  +LSTM        & .859 & .536  & \textbf{.023}  &  .311  \\
  +LSTM+FTM      & \textbf{.863} & \textbf{.547} & \textbf{.023} & \textbf{.328} \\    
  \hline
\end{tabular}
  \vspace{-0.2cm}
  \caption{Results on road scene layout estimation on~\cite{sam:NuScenes18a}. 
  }
  \label{tbl:sub_exp_layout}
\end{table}

\begin{figure*}
    \setlength{\abovecaptionskip}{-0.0cm}    
    \setlength{\belowcaptionskip}{-0.4cm}
    \vspace{-0.2cm}
    \begin{minipage}[h]{0.6\textwidth}
        \centering
        \begingroup
\setlength{\tabcolsep}{1.8pt}
\begin{tabular}{p{3cm}|p{1.7cm}p{1.8cm}p{1cm}p{1.3cm}p{0.9cm}}
  \hline
              &  \multicolumn{5}{c}{KITTI~\cite{sam:Geiger13a}} \\
  \hline
  Method      &  Accu.-Bi. $\uparrow$ & Accu.-Mc. $\uparrow$ & MSE $\downarrow$ & F1.-Bi $\uparrow$ & IOU $\uparrow$  \\
  \hline\hline
  RGB~\cite{sam:Seff16a}          & .811& .778& .230& .176 & .327\\
  RGB~\cite{sam:Seff16a}+D        & .818  & .819  & .154&  .109 & .334\\
  BEV~\cite{Wang_2019_CVPR}      & .820& .797& .141& .324 & .345\\
  BEV~\cite{Wang_2019_CVPR}+GM  & .831& .802& .136& .344 & .357\\
  H-BEV-DA\cite{Wang_2019_CVPR}+GM  & .834 & .831 & \textbf{.134} & .435 & .404\\
  \hline
  \hline
  BEV-C          & .826& .779& .175& .456 & .317\\
  BEV-J        & .840 & .832  & .135 & .458 & \textbf{.416}\\
  BEV-J-O      & .831 & \textbf{.837}  & .142 & .494 & .388\\  
  \hline
  +LSTM        & \textbf{.845}  & .825  & \textbf{.134}  & \textbf{.537}  & .382 \\
  +LSTM+FTM      & \textbf{.842} & \textbf{.841} & \textbf{.134} & \textbf{.534} & \textbf{.405} \\    
  \hline
\end{tabular}
\endgroup


        \makeatletter\def\@captype{table}\makeatother\caption{Main results on road scene layout estimation on KITTI~\cite{sam:Geiger13a}. We can see that: 1) The global (BEV-C, BEV-J) and context (BEV-J-O) information do benefits the single image performance compared to their single image competitors. 2) Introducing LSTM and FTM can further improve the accuracy compared to single image method.}
        \label{tbl:main_exp_layout}
    \end{minipage}
    \hspace{0.2cm}
    \begin{minipage}[h]{0.38\textwidth}
        \centering
        \setlength{\abovecaptionskip}{-0.2cm}    
        \includegraphics[height=0.4\textwidth, trim = 0mm 140mm 170mm 10mm, clip]{figures/teaser.pdf}
        \includegraphics[height=0.4\textwidth, trim = 90mm 140mm 68mm 10mm,clip]{figures/teaser.pdf}
        \makeatletter\def\@captype{figure}\makeatother\caption{Illustrations of baseline model architectures. We denote the upper one as $basic$ and lower one as $blstm$.}
        \label{fig:archs_overview}  
    \end{minipage}
\end{figure*}

\paragraph{Evaluation metrics:}
Since our output space consists of three types of predictions and involves both discrete and continuous variables, we follow the metrics defined in~\cite{Wang_2019_CVPR} and summarize them below.

As for binary variables $\saBin$ (like the existence of crosswalk on the main road or left side-walk) and for multi-class variables $\saMc$ (like the number of lanes on your left), we report the prediction accuracy as $\textrm{Accu.-Bi} = \frac{1}{14} \sum_{k=1}^{14} [p_k = \saBin{}_k]$ and $\textrm{Accu.-Mc} = \frac{1}{2} \sum_{k=1}^{2} [p_k = \saMc{}_k]$. More importantly, since we observe that the binary classes are extremely biased, we also report the F1 score on $\saBin$ which better informs the overall performance. Specifically, $\textrm{F1.-Bi} = \frac{1}{14} \sum_{k=1}^{14} 2 \times \frac{p_k \times r_k}{p_k + r_k}$, where $p_k$ and $r_k$ are the precision and recall rate on $k$-th variable of binary attributes.
For regression variables we use the mean square error (MSE).  

In addition, we also report Intersection-over-Union (IoU) as a overall performance evaluation metric. Specifically, we assume that we can render four-class semantic top-view maps with either the predicted results or the ground-truth annotations. Then we report the average IoU score over all test images. More details on IoU are presented in supplementary.

\subsection{Evaluations on Global and Context Cues}
In this section, we would like to explore the impact of $bev\text{-}final$. To this end, we aim to validate the effectiveness of our proposed context and global information on road layout prediction. We use the basic model without LSTM and FTM, unless otherwise specified. Details for this basic model can be found in $basic$ of Fig.~\ref{fig:archs_overview}. 

\paragraph{Baselines:}
We compare our proposed method with several competing methods presented in~\cite{Wang_2019_CVPR}. The~\textbf{RGB} method takes a single perspective RGB image as input and trains a model~\cite{sam:He16a} that directly outputs predictions for scene attributes. The~\textbf{RGB+D} method shares the same model architecture w.r.t.~\textbf{RGB} but outputs both per-pixel depth map and attribute predictions. Both~\textbf{BEV} and the SOTA~\textbf{H-BEV-DA} output scene attributes with $bev$ as input, which means they are all online methods. The differences lie in: 1) the former uses only $bev$ from real images during training while the latter utilizes additional simulated $bev$; 2) the former uses $basic$ model and the later has a hybrid model as in~\cite{Wang_2019_CVPR}. Note that +GM means a spatial-temporal graphical model is added as post-processing.

%

\paragraph{Our proposals:}
Apart from baselines, we further propose the following methods with different input representations:

\begin{itemize}
\item \textbf{BEV-COLMAP (BEV-C):} 
We propose $bev$-$col$ with global information from videos, as described in Sec.~\ref{sec:input_ind_and_colmap}. Once we get the $bev$-$col$, we feed it to the $basic$ model. 
\item \textbf{BEV-JOINT (BEV-J):} We combine the $bev$ and $bev$-$col$ together and obtain the input for BEV-J. Then we feed the joint input to the $basic$ model.
\item \textbf{BEV-JOINT-OBJ (BEV-J-O):} As described in Sec.~\ref{sec:context}, we further add object information and obtain $bev$-$final$ as the input for BEV-J-O. Again, a $basic$ model is learned.
\end{itemize}

\paragraph{Quantitative results:}
Tab.~\ref{tbl:main_exp_layout} summarizes our main results on KITTI~\cite{sam:Geiger13a}. We have the following observations. Firstly,  comparing BEV-C to BEV, we can see that obtaining global information from 3D point clouds generated with the entire video sequence is better than relying on individual depth and semantic segmentation results, e.g. large improvement in F1 score. This improvement might come from higher quality of $bevinit$, especially in terms of both higher accuracy in regions that are far away from the camera and more consistent results. Second, if we compare BEV-J to BEV-C and BEV, we can see that combining them together can further boost the performance for all attributes, e.g. $1.4\%$, $3.3\%$ and $0.04$ improvement for binary, multiclass and regression tasks. More importantly, we can improve the IoU by around $10\%$. One reason might be that the input from BEV-C focuses more on texture regions that can be both close by and far away while input from BEV keeps more details for all close by regions. They are mutual beneficial in terms of representing scene layout in top-view. Finally, once adding object information (BEV-J-O), we can further boost the F1 score by about $5\%$. In conclusion, we can observe a clear improvement when adding individual components. 

We also report partial results on NuScenes~\cite{sam:NuScenes18a} in Tab.~\ref{tbl:sub_exp_layout} and refer the readers to supplementary for complete ones. As for NuScenes~\cite{sam:NuScenes18a}, we again observe that combining individual and global information provides better results than either one in isolation. Also, adding context information further introduces improvements on almost all tasks.

\paragraph{Ablation study:}
Since the main improvement of the performance boost comes from information of the full video sequence as well as objects, in this section, we conduct ablation studies on: 1) the impact of the quality of top-view maps generated with full video sequences on models, and 2) the impact of the representation of objects on models. Tab.~\ref{tbl:single_ablation_table} summarizes our results, where *-denseC means we change 3D reconstruction parameter of COLMAP to generate much denser $bev$-$col$ as inputs to train our model. And *-C-O means we combine the input from BEV-C with object information and learn to predict the road attributes.

\begin{table}\centering\small
\setlength{\belowcaptionskip}{-0.4cm}
  \begin{tabular}{l|cccc}
  \hline
              &  \multicolumn{4}{c}{KITTI~\cite{sam:Geiger13a}} \\
  \hline
  Method      &  Accu.-Bi. $\uparrow$ & Accu.-Mc. $\uparrow$ & MSE $\downarrow$ & F1.-Bi. $\uparrow$\\
  \hline
  BEV-C          & .826& .779& .175& .456  \\
  \hline
  *-denseC & .826      & .783     & .148 & .496 \\ 
  *-C-O     & .836 & .812 & .146 & .507  \\  
  \hline
  *+LSTM     & .831 & .812 & .136 & .360  \\  
  \hline  
\end{tabular}
  \vspace{-0.2cm}
  \caption{Ablation study on single image based road layout prediction. We abbreviate BEV as $*$ in this table.}
  \label{tbl:single_ablation_table}
\end{table}

\begin{figure*}
    \setlength{\belowcaptionskip}{-0.4cm}
	\centering
	\vspace{-0.4cm}
	\begin{tabular}{ccc}
		\includegraphics[width=0.31\textwidth]{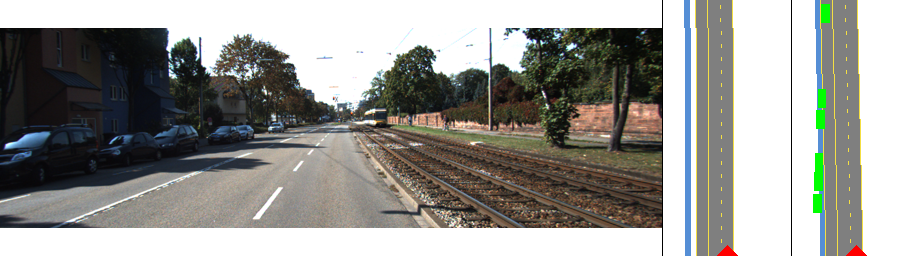} 
		& \includegraphics[width=0.31\textwidth]{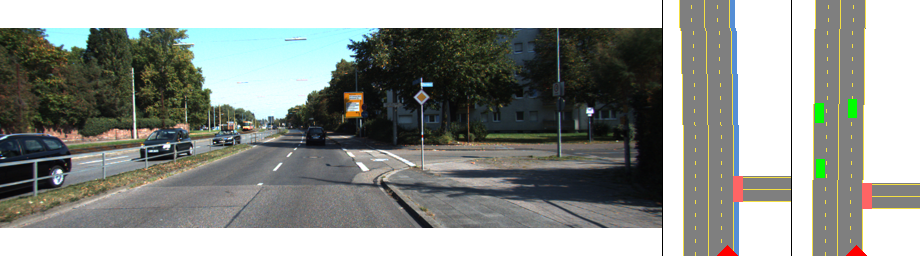}
		&\includegraphics[width=0.31\textwidth]{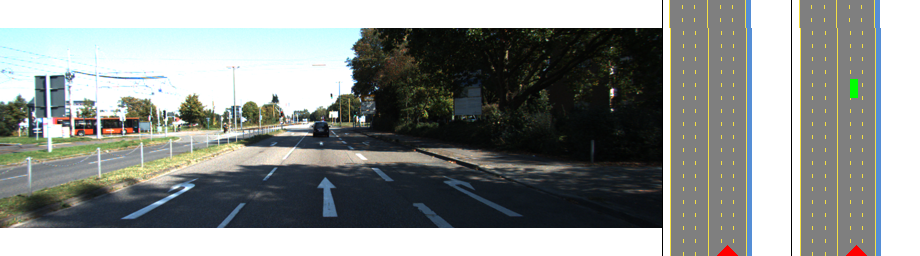} \\
		\includegraphics[width=0.31\textwidth]{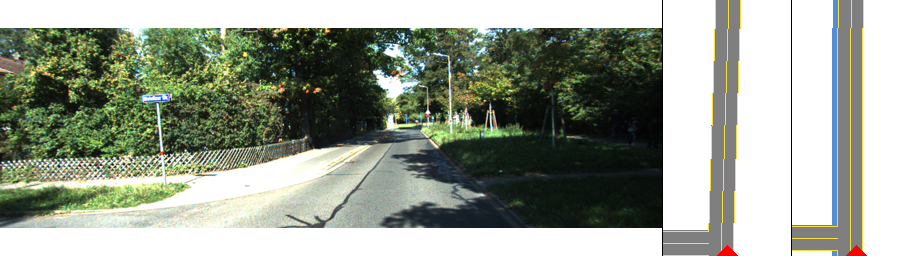}
		&\includegraphics[width=0.31\textwidth]{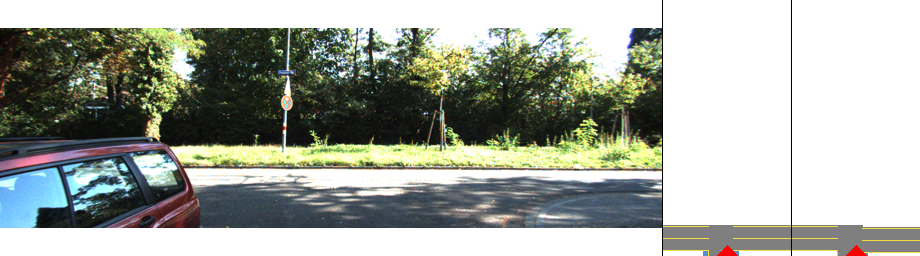} 
		& \includegraphics[width=0.31\textwidth]{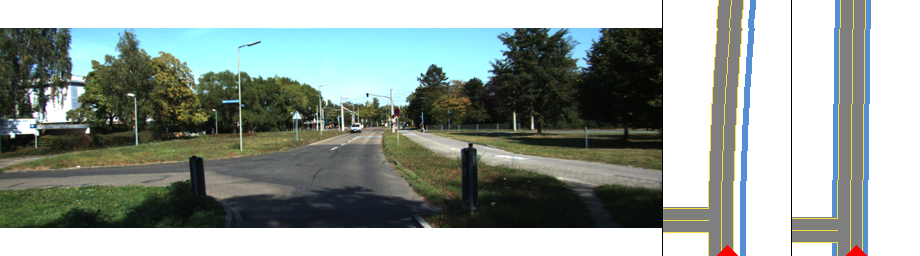} \\
	\end{tabular}
	\vspace{-0.35cm}
	\caption{Qualitative results of full model on individual frames from KITTI.  Each example shows perspective RGB, ground truth and predicted semantic top-view (including the object bounding box as green rectangular), respectively.}
	\label{fig:GM_output_more}
\end{figure*}
 
\begin{figure*}
    \setlength{\belowcaptionskip}{-0.4cm}
	\centering
	\begin{tabular}{ccc}
	\includegraphics[width=0.31\textwidth]{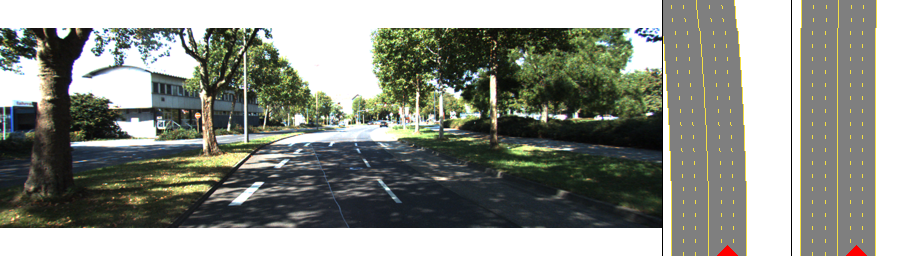} 
	& \includegraphics[width=0.31\textwidth]{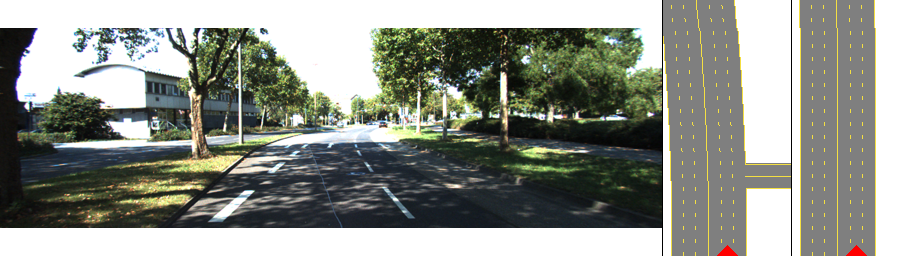} 
	& \includegraphics[width=0.31\textwidth]{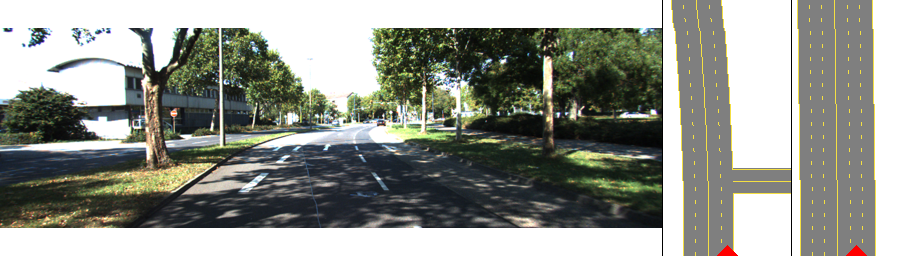} \\
	\includegraphics[width=0.31\textwidth]{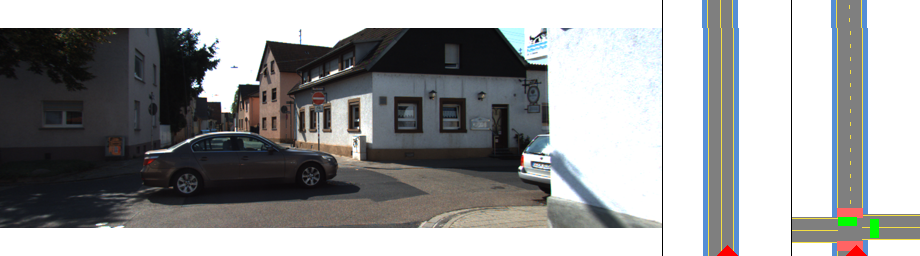} 
	& \includegraphics[width=0.31\textwidth]{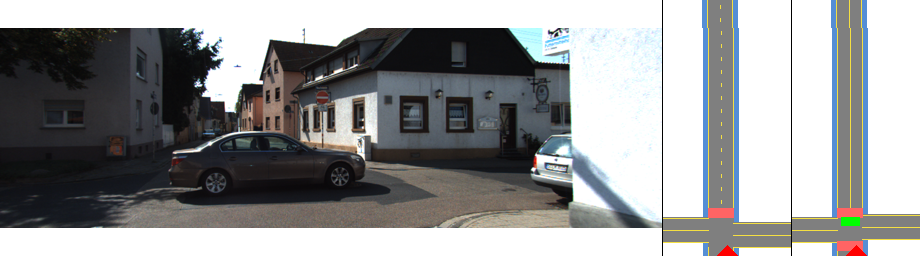} 
	&\includegraphics[width=0.31\textwidth]{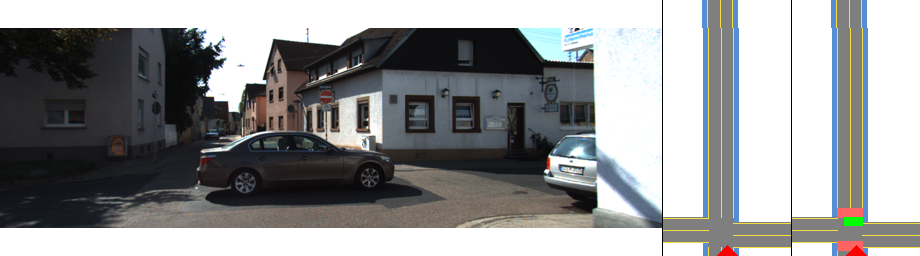} \\
	\end{tabular}
	\vspace{-0.35cm}
	\caption{Qualitative results comparing BEV-J-O and full model in \textbf{consecutive} frames of two example sequences from KITTI.  We visualize for each view the perspective RGB image, prediction from BEV-J-O and from full model, respectively.}
	\label{fig:GM_output}
\end{figure*}

As can be seen in this table, denser $bev$-$col$ can definitely improve the performance for all tasks and measurements, which indicates that there can be more room to boost prediction accuracy by improving the quality of $bev$-$col$. Additionally, further performance boosts can be obtained with object information, which again proves that context information can be very useful in our layout prediction task.

\begin{table}\centering\small
\setlength{\belowcaptionskip}{-0.4cm}

\begin{tabular}{l|cc}
  \hline
              &  \multicolumn{2}{c}{KITTI~\cite{sam:Geiger13a}} \\
  \hline 
  Method      &  seman.$\downarrow$& temp.$\downarrow$ \\ 
  \hline
  BEV~\cite{Wang_2019_CVPR} & 2.65& 3.99\\
  H-BEV-DA~\cite{Wang_2019_CVPR}+GM& 1.77& 1.93\\                  
  \hline
  BEV-J-O & 1.26 & 2.78\\
  \hline
  +LSTM+FTM & 1.26 & 1.96\\
\end{tabular}
  \vspace{-0.3cm}
  \caption{Main results on consistency measurements.}
  \label{tbl:consistency_table}
\end{table}

\paragraph{Qualitative results:}
We show several qualitative results in Fig.~\ref{fig:GM_output_more}. We can see from the examples that our model successfully describes a diverse set of road scenes. Moreover, our predictions are generally consistent w.r.t. object detection results.

\subsection{Evaluations on LSTM and FTM}
We further demonstrate the effectiveness of LSTM and FTM in Tab.~\ref{tbl:main_exp_layout}. More specifically, +LSTM model is illustrated in Fig.~\ref{fig:archs_overview} ($blstm$ part) and +LSTM+FTM is our full model. Both models receive $bev$-$final$ as input.

As can be seen in this table, with the help of LSTM, we can increase performance on the binary and regression tasks. Note that we can further boost the F1 score by at least $4\%$ compared to single image methods. More importantly, our full model, or +LSTM+FTM, provides the best overall performance. As highlighted in the table, it almost always performs the best or second best among all tasks and measurements. Even compared to previous SOTA that is trained with both real and simulated data and includes a spatio-temporal graphical model as post-processing, we can clearly observe that the proposed method provides better results, specifically for F1 score. In conclusion, we demonstrate the effectiveness of FTM and LSTM in terms of improving prediction accuracy. We also report the BEV+LSTM in Tab.~\ref{tbl:single_ablation_table}. We can observe that adding LSTM ($blstm$) can improve the performance even with $bev$ inputs, which demonstrates the effectiveness of temporal information on locally obtained top-view representations. Such trend is also shared on NuScenes in Tab.~\ref{tbl:sub_exp_layout}.

\paragraph{Consistency:}
Since we are working in videos, we also analyze the consistency in predictions. To this end, we utilize the following metrics as suggested in~\cite{Wang_2019_CVPR} and refer the readers to supplementary material for more details:

\begin{itemize}
\item {\em Semantic consistency:} we report the average conflicts in attribute predictions. 
\item {\em Temporal consistency:} we also report the average number of changes in our prediction. 
\end{itemize}

We show in Tab.~\ref{tbl:consistency_table} quantitative results for the temporal as well as semantic consistency metrics defined on the two data sets. Compared to the single image baseline, we can observe significant improvement from the proposed BEV-J-O in consistency for both data sets. Also, if we compare with video-based existing work, combined with the prediction accuracy shown in Tab.\ref{tbl:main_exp_layout}, we can see that the proposed method outperforms previous SOTA, e.g. boost the accuracy while not sacrificing temporal smoothness. Results and our analysis on NuScenes~\cite{sam:NuScenes18a} can be found in the supplementary.  

Finally, we visualize qualitative results of consecutive frames in two test sequences from KITTI in Fig.~\ref{fig:GM_output}. As can be seen, our model successfully enforces temporal smoothness. We can also observe more consistent predictions, \eg, width of side-road and delimiter width, with the help of the LSTM and FTM. Again, our predictions are consistent w.r.t. object detection results.



\section{Conclusion}
\label{sec:conclusion}
In this work, we present a scene understanding framework to estimate the parametric road layout in each view in a video. In constrast to the case with only single image input, we propose to make use of the temporal information in videos by leveraging LSTM/FTM model, context information as well as global 3D reconstruction, altogether leading to very promising results.  

{\small
\bibliographystyle{ieee}
\bibliography{myshortstrings,egbib}

\begin{thebibliography}{10}\itemsep=-1pt

\bibitem{sam:Armeni16a}
Iro Armeni, Ozan Sener, Amir~R. Zamir, Helen Jiang, Ioannis Brilakis, Martin
  Fischer, and Silvio Savarese.
\newblock {3D Semantic Parsing of Large-Scale Indoor Spaces}.
\newblock In {\em CVPR}, 2016.

\bibitem{sam:RotaBulo18a}
Samuel~Rota Bul\`{o}, Lorenzo Porzi, and Peter Kontschieder.
\newblock {In-Place Activated BatchNorm for Memory-Optimized Training of DNNs}.
\newblock In {\em CVPR}, 2018.

\bibitem{sam:Chen18a}
Liang-Chieh Chen, Yukun Zhu, George Papandreou, Florian Schroff, and Hartwig
  Adam.
\newblock {Encoder-Decoder with Atrous Separable Convolution for Semantic Image
  Segmentation}.
\newblock In {\em ECCV}, 2018.

\bibitem{donahue2015long}
Jeffrey Donahue, Lisa Anne~Hendricks, Sergio Guadarrama, Marcus Rohrbach,
  Subhashini Venugopalan, Kate Saenko, and Trevor Darrell.
\newblock Long-term recurrent convolutional networks for visual recognition and
  description.
\newblock In {\em Proceedings of the IEEE conference on computer vision and
  pattern recognition}, pages 2625--2634, 2015.

\bibitem{feichtenhofer2017spatiotemporal}
Christoph Feichtenhofer, Axel Pinz, and Richard~P Wildes.
\newblock Spatiotemporal multiplier networks for video action recognition.
\newblock In {\em Proceedings of the IEEE conference on computer vision and
  pattern recognition}, pages 4768--4777, 2017.

\bibitem{fischer2015flownet}
Philipp Fischer, Alexey Dosovitskiy, Eddy Ilg, Philip H{\"a}usser, Caner
  Haz{\i}rba{\c{s}}, Vladimir Golkov, Patrick Van~der Smagt, Daniel Cremers,
  and Thomas Brox.
\newblock Flownet: Learning optical flow with convolutional networks.
\newblock {\em arXiv preprint arXiv:1504.06852}, 2015.

\bibitem{sam:Geiger14a}
Andreas Geiger, Martin Lauer, Christian Wojek, Christoph Stiller, and Raquel
  Urtasun.
\newblock {3D Traffic Scene Understanding from Movable Platforms}.
\newblock {\em PAMI}, 2014.

\bibitem{sam:Geiger13a}
Andreas Geiger, Philip Lenz, Christoph Stiller, and Raquel Urtasun.
\newblock {Vision meets Robotics: The KITTI Dataset}.
\newblock {\em International Journal of Robotics Research (IJRR)}, 2013.

\bibitem{sam:Godard17a}
Cl\'{e}ment Godard, Oisin~Mac Aodha, and Gabriel~J. Brostow.
\newblock {Unsupervised Monocular Depth Estimation with Left-Right
  Consistency}.
\newblock In {\em CVPR}, 2017.

\bibitem{sam:Guo12a}
Ruiqi Guo and Derek Hoiem.
\newblock {Beyond the line of sight: labeling the underlying surfaces}.
\newblock In {\em ECCV}, 2012.

\bibitem{sam:Gupta17a}
Saurabh Gupta, James Davidson, Sergey Levine, Rahul Sukthankar, and Jitendra
  Malik.
\newblock {Cognitive Mapping and Planning for Visual Navigation}.
\newblock In {\em CVPR}, 2017.

\bibitem{sam:He16a}
Kaiming He, Xiangyu Zhang, Shaoqing Ren, and Jian Sun.
\newblock {Deep Residual Learning for Image Recognition}.
\newblock In {\em CVPR}, 2016.

\bibitem{hochreiter1997long}
Sepp Hochreiter and J{\"u}rgen Schmidhuber.
\newblock Long short-term memory.
\newblock {\em Neural computation}, 9(8):1735--1780, 1997.

\bibitem{sam:Jaderberg15a}
Max Jaderberg, Karen Simonyan, Andrew Zisserman, and Koray Kavukcuoglu.
\newblock {Spatial Transformer Networks}.
\newblock In {\em NIPS}, 2015.

\bibitem{sam:Kunze18a}
Lars Kunze, Tom Bruls, Tarlan Suleymanov, and Paul Newman.
\newblock {Reading between the Lanes: Road Layout Reconstruction from Partially
  Segmented Scenes}.
\newblock In {\em International Conference on Intelligent Transportation
  Systems (ITSC)}, 2018.

\bibitem{sam:Laina16a}
Iro Laina, Vasileios~Belagiannis Christian~Rupprecht, Federico Tombari, and
  Nassir Navab.
\newblock {Deeper Depth Prediction with Fully Convolutional Residual Networks}.
\newblock In {\em 3DV}, 2016.

\bibitem{li2019gs3d}
Buyu Li, Wanli Ouyang, Lu Sheng, Xingyu Zeng, and Xiaogang Wang.
\newblock Gs3d: An efficient 3d object detection framework for autonomous
  driving.
\newblock In {\em Proceedings of the IEEE Conference on Computer Vision and
  Pattern Recognition}, pages 1019--1028, 2019.

\bibitem{li2019stereo}
Peiliang Li, Xiaozhi Chen, and Shaojie Shen.
\newblock Stereo r-cnn based 3d object detection for autonomous driving.
\newblock In {\em Proceedings of the IEEE Conference on Computer Vision and
  Pattern Recognition}, pages 7644--7652, 2019.

\bibitem{liu2015multiclass}
Buyu Liu and Xuming He.
\newblock Multiclass semantic video segmentation with object-level active
  inference.
\newblock In {\em Proceedings of the IEEE conference on computer vision and
  pattern recognition}, pages 4286--4294, 2015.

\bibitem{sam:Liu15a}
Chenxi Liu, Alexander~G. Schwing, Kaustav Kundu, Raquel Urtasun, and Sanja
  Fidler.
\newblock {Rent3D: Floor-Plan Priors for Monocular Layout Estimation}.
\newblock In {\em CVPR}, 2015.

\bibitem{sam:Mattyus16a}
Gell\'{e}rt M\'{a}ttyus, Shenlong Wang, Sanja Fidler, and Raquel Urtasun.
\newblock {HD Maps: Fine-grained Road Segmentation by Parsing Ground and Aerial
  Images}.
\newblock In {\em CVPR}, 2016.

\bibitem{sam:NuScenes18a}
{NuTonomy}.
\newblock {The NuScenes data set}.
\newblock \url{ https://www.nuscenes.org }, 2018.

\bibitem{schonberger2016structure}
Johannes~L Schonberger and Jan-Michael Frahm.
\newblock Structure-from-motion revisited.
\newblock In {\em Proceedings of the IEEE Conference on Computer Vision and
  Pattern Recognition}, pages 4104--4113, 2016.

\bibitem{schonberger2016pixelwise}
Johannes~L Sch{\"o}nberger, Enliang Zheng, Jan-Michael Frahm, and Marc
  Pollefeys.
\newblock Pixelwise view selection for unstructured multi-view stereo.
\newblock In {\em European Conference on Computer Vision}, pages 501--518.
  Springer, 2016.

\bibitem{sam:Schulter18a}
Samuel Schulter, Menghua Zhai, Nathan Jacobs, and Manmohan Chandraker.
\newblock {Learning to Look around Objects for Top-View Representations of
  Outdoor Scenes}.
\newblock In {\em ECCV}, 2018.

\bibitem{schuster1997bidirectional}
Mike Schuster and Kuldip~K Paliwal.
\newblock Bidirectional recurrent neural networks.
\newblock {\em IEEE Transactions on Signal Processing}, 45(11):2673--2681,
  1997.

\bibitem{sam:Seff16a}
Ari Seff and Jianxiong Xiao.
\newblock {Learning from Maps: Visual Common Sense for Autonomous Driving}.
\newblock {\em arXiv:1611.08583}, 2016.

\bibitem{sam:Sengupta12a}
Sunando Sengupta, Paul Sturgess, \`Lubor Ladick\'{y}, and Philip H.~S. Torr.
\newblock {Automatic Dense Visual Semantic Mapping from Street-Level Imagery}.
\newblock In {\em IROS}, 2012.

\bibitem{simonyan2014two}
Karen Simonyan and Andrew Zisserman.
\newblock Two-stream convolutional networks for action recognition in videos.
\newblock In {\em Advances in neural information processing systems}, pages
  568--576, 2014.

\bibitem{sam:Song17a}
Shuran Song, Fisher Yu, Andy Zeng, Angel~X. Chang, Manolis Savva, and Thomas
  Funkhouser.
\newblock {Semantic Scene Completion from a Single Depth Image}.
\newblock In {\em CVPR}, 2017.

\bibitem{sam:Song18a}
Shuran Song, Andy Zeng, Angel~X. Chang, Manolis Savva, Silvio Savarese, and
  Thomas Funkhouser.
\newblock {Im2Pano3D: Extrapolating 360 Structure and Semantics Beyond the
  Field of View}.
\newblock In {\em CVPR}, 2018.

\bibitem{srivastava2015unsupervised}
Nitish Srivastava, Elman Mansimov, and Ruslan Salakhudinov.
\newblock Unsupervised learning of video representations using lstms.
\newblock In {\em International conference on machine learning}, pages
  843--852, 2015.

\bibitem{sun2010secrets}
Deqing Sun, Stefan Roth, and Michael~J Black.
\newblock Secrets of optical flow estimation and their principles.
\newblock In {\em 2010 IEEE computer society conference on computer vision and
  pattern recognition}, pages 2432--2439. IEEE, 2010.

\bibitem{Tighe_2014_CVPR}
Joseph Tighe, Marc Niethammer, and Svetlana Lazebnik.
\newblock {Scene Parsing with Object Instances and Occlusion Ordering}.
\newblock In {\em CVPR}, June 2014.

\bibitem{sam:Tulsiani18a}
Shubham Tulsiani, Richard Tucker, and Noah Snavely.
\newblock {Layer-structured 3D Scene Inference via View Synthesis}.
\newblock In {\em ECCV}, 2018.

\bibitem{vu2018memory}
Tuan-Hung Vu, Wongun Choi, Samuel Schulter, and Manmohan Chandraker.
\newblock Memory warps for learning long-term online video representations.
\newblock {\em arXiv preprint arXiv:1803.10861}, 2018.

\bibitem{Wang_2019_CVPRworkshop}
Ziyan Wang, Buyu Liu, Samuel Schulter, and Manmohan Chandraker.
\newblock A dataset for high-level 3d scene understanding of complex road
  scenes in the top-view.
\newblock In {\em The IEEE Conference on Computer Vision and Pattern
  Recognition (CVPR) Workshop}, June 2019.

\bibitem{Wang_2019_CVPR}
Ziyan Wang, Buyu Liu, Samuel Schulter, and Manmohan Chandraker.
\newblock A parametric top-view representation of complex road scenes.
\newblock In {\em The IEEE Conference on Computer Vision and Pattern
  Recognition (CVPR)}, June 2019.

\bibitem{xingjian2015convolutional}
SHI Xingjian, Zhourong Chen, Hao Wang, Dit-Yan Yeung, Wai-Kin Wong, and
  Wang-chun Woo.
\newblock Convolutional lstm network: A machine learning approach for
  precipitation nowcasting.
\newblock In {\em Advances in neural information processing systems}, pages
  802--810, 2015.

\bibitem{sam:Xu18a}
Dan Xu, Wei Wang, Hao Tang, Hong Liu, Nicu Sebe, and Elisa Ricci.
\newblock {Structured Attention Guided Convolutional Neural Fields for
  Monocular Depth Estimation}.
\newblock In {\em CVPR}, 2018.

\bibitem{sam:Zhao17a}
Hengshuang Zhao, Jianping Shi, Xiaojuan Qi, Xiaogang Wang, and Jiaya Jia.
\newblock {Pyramid Scene Parsing Network}.
\newblock In {\em CVPR}, 2017.

\bibitem{zhu2017flow}
Xizhou Zhu, Yujie Wang, Jifeng Dai, Lu Yuan, and Yichen Wei.
\newblock Flow-guided feature aggregation for video object detection.
\newblock In {\em Proceedings of the IEEE International Conference on Computer
  Vision}, pages 408--417, 2017.

\bibitem{zhu2017deep}
Xizhou Zhu, Yuwen Xiong, Jifeng Dai, Lu Yuan, and Yichen Wei.
\newblock Deep feature flow for video recognition.
\newblock In {\em Proceedings of the IEEE Conference on Computer Vision and
  Pattern Recognition}, pages 2349--2358, 2017.

\end{thebibliography}
}

\end{document}